\documentclass[letterpaper]{article}
\usepackage{graphicx}
\usepackage{aaai}
\usepackage{times}
\usepackage{helvet}
\usepackage{courier}
\usepackage[utf8]{inputenc}
\usepackage{amsmath}
\usepackage{bbm}
\usepackage{footnote}
\usepackage{multirow}
\usepackage{comment}
\usepackage{booktabs}
\usepackage{float}
\usepackage{array}
\usepackage{url}
\makesavenoteenv{tabular}
\makesavenoteenv{table}

\newcolumntype{M}[1]{>{\centering\arraybackslash}m{#1}}

\usepackage{todonotes}

\newcolumntype{P}[1]{>{\centering\arraybackslash}p{#1}}

\begin{document}
\title{Boosting Image Recognition with Non-differentiable Constraints}
\author{Xuan Li,\textsuperscript{1}
Yuchen Lu,\textsuperscript{2}
Peng Xu,\textsuperscript{3}
Jizong Peng,\textsuperscript{4}
Christian Desrosiers,\textsuperscript{4}
Xue Liu\textsuperscript{1}
\\
\textsuperscript{1}{McGill University}\\
\textsuperscript{2}{Universite de Montreal}\\
\textsuperscript{3}{Polytechnique Montreal}\\
\textsuperscript{4}{ETS Montreal}\\
xuan.li2@mcgill.ca,
yuchen.lu@umontreal.ca,
peng.xu@polymtl.ca,
jizong.peng.1@etsmtl.net,
christian.desrosiers@etsmtl.ca,
xue.liu@mcgill.ca}
\maketitle

\begin{abstract}
\begin{quote}
In this paper, we study the problem of image recognition with non-differentiable constraints. A lot of real-life recognition applications require a rich output structure with deterministic constraints that are discrete or modeled by a non-differentiable function. A prime example is recognizing digit sequences, which are restricted by such rules (e.g., \textit{container code detection}, \textit{social insurance number recognition}, etc.). We investigate the usefulness of adding non-differentiable constraints in learning for the task of digit sequence recognition. Toward this goal, we synthesize six different datasets from MNIST and Cropped SVHN, with three discrete rules inspired by real-life protocols. To deal with the non-differentiability of these rules, we propose a reinforcement learning approach based on the policy gradient method. We find that incorporating this rule-based reinforcement can effectively increase the accuracy for all datasets and provide a good inductive bias which improves the model even with limited data. On one of the datasets, MNIST\_Rule2, models trained with rule-based reinforcement increase the accuracy by 4.7\% for 2000 samples and 23.6\% for 500 samples. We further test our model against synthesized adversarial examples, e.g., blocking out digits, and observe that adding our rule-based reinforcement increases the model robustness with a relatively smaller performance drop. 
\end{quote}
\end{abstract}

\section{Introduction}
There has been a rising interest in applying deep learning to the problem of Optical Character Recognition (OCR), with numerous datasets focusing on this task \cite{wang2011end,karatzas2015icdar,veit2016coco,netzer2011reading}. However, most of these datasets focus on the perception part of the problem with various kinds of realistic images. Nevertheless, domain-specific rules play an important part in various OCR tasks. For example, Social Insurance Number (SIN) uses Luhn algorithm \cite{luhn1960computer} to generate qualified digits, suggesting that a successful SIN recognition model should take into account the underlying verification rule. We argue that, by equipping our model with the ability of reasoning over these specific rules, the algorithm can perform better in many real-life OCR applications. To fill in the gap, we create six datasets by combining images from MNIST \cite{lecun1998gradient} and Cropped SVHN \cite{netzer2011reading}, respectively, using three different rules, inspired by real-life scenarios. A sample of our dataset can be found in Figure \ref{fig:datasets_mnist} and Figure \ref{fig:datasets_svhn}.

\begin{figure}[h]
    \includegraphics[scale=0.5]{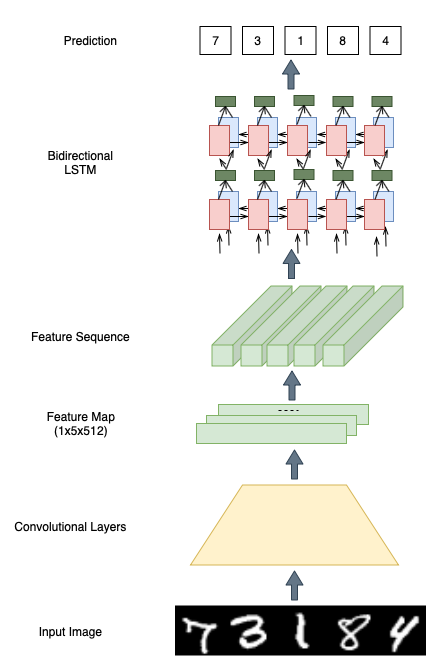}
    \centering
    \caption{The Network Architecture}
    \label{fig:model}
\end{figure}

To test whether existing models can perform well under these datasets, we first implement the Convolutonal Recurrent Neural Network (CRNN) \cite{shi2016end}. The model first uses convolutional layers to extract features, which are then sent to a Bidirectional LSTM. Different from the original work, since we have a fixed number of characters to recognize, we only produce a single prediction for each character and apply cross-entropy loss. An illustration of our architecture can be found in Figure \ref{fig:model}. To incorporate non-differentiable rules into training, we add an additional term in the loss function measuring the expected rule reward which is optimized by a policy gradient algorithm. We further investigate the effect of different hyper-parameters and evaluate how the added rule can help the model learn with reduced data, which is a common problem in real-life scenarios. 
Our contributions can be summarized as follows:
\begin{itemize}
    \item We propose a suite of rule-based OCR benchmark datasets with three different rules inspired by real-life applications;
    
    \item We propose a general method based on reinforcement learning (RL) to include non-differentiable constraints in the learning process of image recognition problems;
    
    \item We conduct an in-depth evaluation of our method on the proposed datasets. Our results show that the proposed rule-based reinforcement can help improve accuracy across different kinds of images and rules. We further demonstrate the effect of our method under different hyper-parameter settings, as well as its effectiveness in challenging learning scenarios involving limited data or missing/indistinguishable digits. 
\end{itemize}

\section{Related Works}

Numerous datasets have been proposed for OCR and digit recognition tasks. MNIST is one of the classic datasets for such problems, largely popularized by the successful application of deep convolutional neural networks (CNNs) \cite{lecun1998gradient}. The Street View House Numbers (SVHN) is another early dataset with more real-life images \cite{goodfellow2013multi}. Following the spirit of large-scale text recognition in the wild, ICDAR robust reading competition was proposed in 2013 \cite{karatzas2013icdar} and updated in the following years \cite{karatzas2015icdar}. It has become one of the standard benchmarks due to its high image quality. Following its success, other large-scale scene text datasets like SVT \cite{svtdata} and COCO-Text \cite{veit2016coco} were created. So far, these datasets have mainly focused on the perception part of the problem. In contrast, the main objective of our dataset is to evaluate whether adding discrete rules while training the model can improve its recognition accuracy in testing.

There have been efforts on the application side to combine OCR with rule matching. Early works in container code identification added the verification rule during the inference phase to improve accuracy \cite{mei2016novel}. Our method differs from this previous work since it includes rules in the training phase by directly optimizing for it.

Our task is also related to numeric reasoning, which has been an important topic in deep learning \cite{dehaene2011number}. Although there has been a series of work on counting objects in images \cite{dijkstra2018centroidnet} or in a sequential manner \cite{fang2018can}, the harder problem on arithmetic or number theory remain largely untouched. Our work can be thought of as a step toward the direction of marrying arithmetics with computer vision.

Our proposed method is related to structured text generation in natural language processing (NLP). The seminal work of \citeauthor{ranzato2015sequence} (\citeyear{ranzato2015sequence}) proposes to use RL for optimizing sequence-level metrics like BLEU and apply it to neural machine translation. There have been similar techniques in other tasks like abstractive summarization \cite{dong2018banditsum} and goal-oriented dialogue \cite{shah2016interactive}. Although we do not have a sequential generation problem, our proposed rule-based reinforcement method is similar to these works. 

\section{Methodology}

Suppose we have a dataset $\{x_i, y_i\}_{i=1}^N$. We formally define a rule as a function $r$ mapping $y$ in output space to a real number. Suppose our recognition system is a neural network which models the conditional distribution $p(y|x;\theta)$, where $\theta$ are the parameters. The objective to maximize for each example is defined as
\begin{equation}\label{eqn:objective}
(1-\alpha)\log p(y_i | x_i;\theta) + \alpha E_{\hat y \sim p(\cdot|x_i;\theta)}[r(\hat y)]
\end{equation}
The first term is the likelihood term, while the second term is the expected reward under the current model. The weight $\alpha$ is used to balance two objectives. By employing the score function trick of the reinforce algorithm \cite{williams1992simple}, the gradient of Eqn.~(\ref{eqn:objective}) can be expressed as
\begin{align}
&\quad (1-\alpha)\nabla_\theta\log p(y_i | x_i;\theta) + \alpha \int \nabla_\theta p(\hat y|x_i;\theta)r(\hat y) d\hat y \nonumber\\
&=(1-\alpha)\nabla_\theta\log p(y_i | x_i;\theta) \nonumber\\   
&\qquad +\alpha \int p(\hat y|x_i;\theta) \nabla_\theta \log p(\hat y|x_i;\theta) r(\hat y) d\hat y \nonumber\\    
&=(1-\alpha)\nabla_\theta \log p(y_i | x_i;\theta) \nonumber\\ 
&\qquad + \alpha E_{\hat y \sim p(\cdot|x_i;\theta)}[r(\hat y) \nabla_\theta \log p(\hat y | x_i;\theta)]
\end{align}
In practice, we approximate the second expectation via sampling $\hat y_j \sim p(\cdot | x_i;\theta)$, and thus our final gradient is
\begin{equation}\label{eqn:gradient}
(1-\alpha)\nabla_\theta \log p(y_i | x_i;\theta) + \alpha \sum_{j=1}^M r(\hat y_j) \nabla_\theta \log p(\hat y_j | x_i;\theta)    
\end{equation}
where $M$ is the number of samples. The second term can also be viewed as policy gradients with a one-step MDP. Since our neural network outputs the logits for each character, we can perform categorical sampling.

\section{Dataset}

Due to the specific characteristics of our hypothesis, we construct six synthesized benchmark datasets using both MNIST \cite{lecun1998gradient} and Cropped SVHN \cite{netzer2011reading} datasets. In our new datasets, each image has 5 digits which are horizontally concatenated by 5 single images. We design three discrete rules to create the datasets and evaluate our method. 
\subsection{Rule\,1}
\begin{figure}[ht!]
\includegraphics[width=0.7\linewidth]{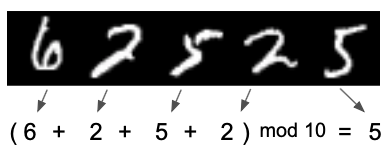}
\centering
\caption{Illustration of Rule\,1.}
\label{fig:datasets_rule1}
\end{figure}

This rule requires the fifth digit to be the remainder of the sum of the first 4 digits with respect to modulus 10 (Fig. \ref{fig:datasets_rule1}).
\par
To synthesize the dataset following this rule, we fist randomly sample numbers from 0 to 9 for the first 4 digits. Then, for each digit, we randomly sample an image from the original dataset of that number directory. The 5-th digit is obtained by applying Rule\,1. Then, we sample an image from the corresponding directory. Each of the synthesized image is comprised of the five horizontally-concatenated images, where images are sampled from MNIST or Cropped SVHN, respectively. The final synthesized image and Rule\,1 can be seen in Fig. \ref{fig:datasets_rule1}). 

\subsection{Rule\,2}

\begin{figure}[ht!]
\includegraphics[width=0.85\linewidth]{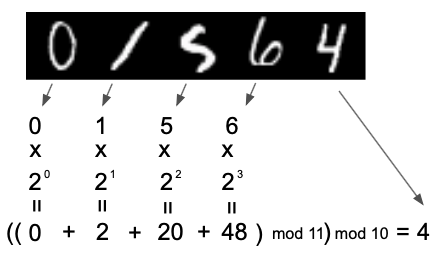}
\centering
\caption{Illustration of Rule\,2.}
\label{fig:datasets_rule2}
\end{figure}

This second rule comes from the ISO 6346 protocol \cite{ISO6346} used for container code recognition. To verify if a 5-digit sequence satisfies this rule, each of the leading 4 digits is multiplied by $2^\mathrm{pos}$ where $\mathrm{pos}=0,1,2,3$. Then we sum up the resulting numbers, and find its remainder respect to modulus 11. If the remainder is 10, then the last digit should be 0, otherwise it should be the remainder. The generation of the dataset satisfying this rule is similar to Rule\,1, once each digit is chosen, we randomly sample images from the corresponding image folder (see Fig.~\ref{fig:datasets_rule2}). 

\subsection{Rule\,3}

\begin{figure}[ht!]
\includegraphics[width=0.85\linewidth]{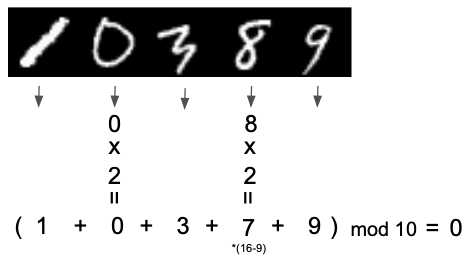}
\centering
\caption{illustration of Rule\,3.}
\label{fig:datasets_rule3}
\end{figure}

The third rule is from the Luhn algorithm \cite{luhn1960computer} which is commonly used to verify Social Insurance Numbers or credit card numbers. To verify this rule for a sequence, the even-indexed numbers are first multiplied by 2, subtracting 9 if the product is greater than 9. The resulting numbers are then summed up along with the original odd-indexed numbers. According to the rule, the final result is required to be divisible by 10 (Figure \ref{fig:datasets_rule3}). To generate the dataset satisfying this rule, a random number between 1000 and 9999 is randomly sampled, and then used as the first 4 digits in the string. To get the qualified 5-th digit, a Luhn generator is used \cite{McLoughlin2015}. The generator takes the 4-digit string as input, and outputs a number that can be appended to the 5-th digit, making the whole 5-digit string Luhn string. The generation of the dataset from the string is similar to the previous rules. We randomly sample images from corresponding image folder.

\par
\begin{figure}[ht!]
\includegraphics[width=\linewidth]{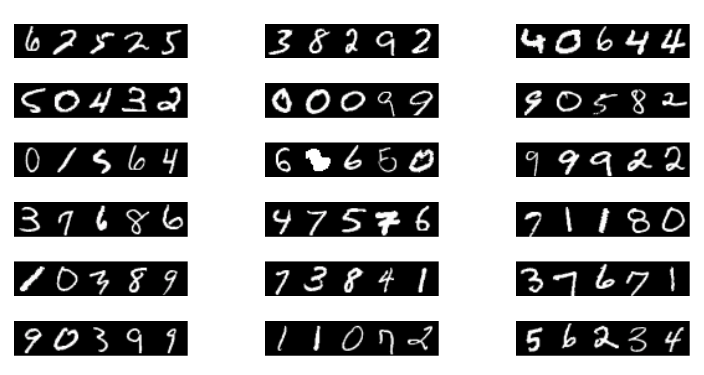}
\centering
\caption{Samples of the synthesized images from MNIST. Top two rows are from Rule\,1, middle two rows are from Rule\,2, bottom two rows are from Rule\,3}
\label{fig:datasets_mnist}
\end{figure}
\begin{figure}[ht!]
\includegraphics[width=\linewidth]{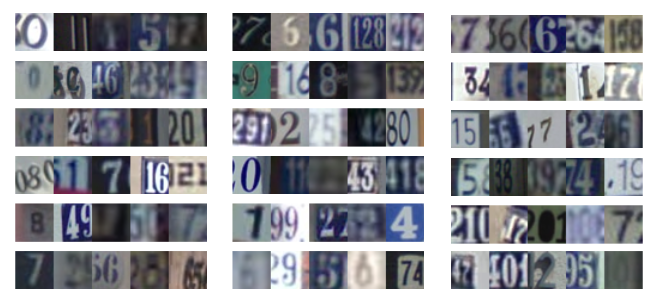}
\centering
\caption{Samples of the synthesized images from cropped SVHN. Top two rows are from Rule\,1, middle two rows are from Rule\,2, bottom two rows are from Rule\,3}
\label{fig:datasets_svhn}
\end{figure}

As shown in Table \ref{tab:all_datasets}, we end up with a total of 6 datasets. Each dataset consists of a total of 3000 images: 2000 randomly sampled images for training, 500 for validation, and 500 for testing. Examples of synthesized images are displayed in Fig.~\ref{fig:datasets_mnist} and Fig.~\ref{fig:datasets_svhn}.

\begin{table}[ht!]
    \centering
    \caption{Synthesized datasets}
    \label{tab:all_datasets}
\begin{tabular}{ P{1.0cm}|P{2.5cm}|P{2.5cm}  }
 \midrule
  & MNIST & Cropped SVHN \\
 \midrule
 Rule\,1 & MNIST\_Rule\,1 & SVHN\_Rule\,1 \\ 
 Rule\,2 & MNIST\_Rule2 & SVHN\_Rule2 \\ 
 Rule\,3 & MNIST\_Rule3 & SVHN\_Rule3 \\ 
 \midrule
\end{tabular}
\end{table}

We conduct our experiments on the six datasets with different combinations of three rules and images from two different sources. The datasets and models will be released upon acceptance.

\section{Experiments}
\subsection{Setup}
We adopt CRNN \cite{shi2016end} as our baseline model to perform sequential recognition. Contrary to other traditional OCR evaluation metric focused on per character accuracy, we argue that the rule-based OCR should have a more strict assessment. We consider sequence-level accuracy:
\begin{equation}
    \text{Accuracy}(\hat{y}_i, y_i)=\left\{
    \begin{array}{@{}ll@{}}
    1, & \text{if}\ \hat{y}^j_i=y^j_i \text{ for j = 0 to 4} \\
    0, & \text{otherwise}
    \end{array}\right.
\end{equation}
where $\hat{y}^j_i$ is the $j$-th digit of $i$-th image, and $y^i$ is the ground truth of $i$-th image. The prediction is correct if and only if all 5 digits are correctly recognized. During all experiments, we simply take the argmax of logits as the final predictions. We leave the investigation of more advanced decoding techniques for future work.

In each training, we resize input images to 28$\times$112. We use Eqn. (\ref{eqn:gradient}) with a batch size of 100 to compute the gradients and employ Adam \cite{kingma2014adam} to train networks for a total of 200 epochs. The learning rate is set to $1 \times 10^{-3}$ and divided by 10 every 60 epochs. We use Pytorch \cite{paszke2017automatic} for all experiments. We choose $\alpha=0.05$ and $M=10000$ if without other specification.

\subsection{The Effect of $\alpha$}
Parameter $\alpha$ controls the relative weight between the cross-entropy loss and the rule-based reinforcement term. We test the following values: 0, 0.001, 0.005, 0.01, 0.05, 0.1 and 0.5. When $\alpha=0$, the model only uses the cross-entropy loss. Additionally, we test two strategies to update this parameter dynamically, called adaptive ascending ($AA$) and adaptive descending ($AD$). For adaptive ascending, we use
\begin{equation}
    AA = \exp\Big(1-\frac{T}{i+1}\Big)
    \label{adapative_weight}
\end{equation}
where $i$ is current epoch number, and $T$ is the total number of epochs. On the other hand, for adaptive descending we set 
$$AD = 1 - AA$$
\par
The results can be found in Table \ref{tab:experiment1}. We first observe that adding rule-based reinforcement improves the performance across all three datasets consistently, showing the effectiveness of our proposed method. When increasing $\alpha$, we find that the model's performance first increases but then decreases. This can be explained by the fact that our rule constraint does not dependent on the label, thus the model may only focus on producing well-defined sequence instead of recognizing actual digits. The optimal value of $\alpha$ can different in each dataset, so in practice it should be tuned to get the best performance. We also find that the adaptive ascending scheme ($AA$) outperforms the baseline model while the adaptive descending scheme ($AD$) decreases accuracy. We argue that it is more beneficial to let the cross-entropy loss be dominant in the beginning and only use rule-base reinforcement to improve the model in the later stage.

\begin{table}[ht!]
    \centering
    \caption{Accuracy (\%) on different $\alpha$}
    \label{tab:experiment1}
\begin{tabular}{ P{0.8cm}P{1.9cm}P{1.9cm}P{1.9cm}}
 \toprule
 $\alpha$ & MNIST\_Rule\,1 & MNIST\_Rule2 & MNIST\_Rule3  \\
 \midrule
0     & 89.4         & 89.6          & 90.4  \\ 
0.001 & 88.8         & 90.7          & 89.0  \\ 
0.005 & \textbf{91.0} & 93.4          & 90.8  \\ 
0.01  & 91.0        & 91.4          & 91.4  \\ 
0.05  & 90.0         & 92.0          & 91.8  \\
0.1   & 90.0         & \textbf{94.3} & 93.2 \\
0.5   & 72.4         & 0             & 92.5 \\ 
\midrule
$AA$ & 90.0 & 92.6 & \textbf{94.4} \\
$AD$ & 42.2 & 0 & 0  \\
\toprule
 &  & & \\
\toprule
 $\alpha$ & SVHN\_Rule\,1 & SVHN\_Rule2 & SVHN\_Rule3 \\
 \midrule
 0 & 22.1  & 31.4 & 28.2 \\
 0.001 & 23.3  & 31.2 & 28.0 \\
 0.005 & 22.6  & 31.5 & 34.8 \\
 0.01 & 25.2  & 32.2 & 34.2 \\
 0.05 & 24.6  & 35.0 & 35.8 \\
 0.1 & \textbf{25.9}  & \textbf{38.6} & \textbf{39.8} \\
 0.5 & 24.3  & 0 & 23.2 \\
 \midrule
 $AA$ & 22.2 & 35.0 & 33.2 \\
 $AD$ & 0 & 0 & 0 \\
\bottomrule
\end{tabular}
\end{table}

\subsection{Learning with Limited Data}
This experiment explores the effect of training dataset size with values from $N=500$ to 2000. Results of this experiments are provided in Table \ref{tab:experiment2} along with relative accuracy gain. In most cases, adding rule-based reinforcement improves the performance regardless of the dataset size. It can be noticed that, in MNIST\_Rule2 and MNIST\_Rule3, the relative accuracy gain is larger when data is most limited (i.e., 500 samples), with 77.6\% relative gain in MNIST\_Rule2 and 11.5\% in MNIST\_Rule3. These results suggest that, in some scenarios, enforcing the rule introduces a good inductive bias which helps the model learn with limited data. However, the same observation is made for the SVHN based dataset. We hypothesize that SVHN uses real-life images so it requires more data to have reasonable performance.

\begin{table*}[ht!]
    \centering
    \caption{Accuracy (\%) for different dataset sizes $N$.}
    \label{tab:experiment2}
\begin{tabular}{ M{1.3cm}M{1.3cm}M{1.3cm}M{1.3cm}M{0.1cm}M{1.3cm}M{1.3cm}M{1.3cm}M{0.1cm}M{1.3cm}M{1.3cm}M{1.3cm}}
 \toprule
 \multirow{1}{*}{} & \multicolumn{3}{c}{MNIST\_Rule\,1} & & \multicolumn{3}{c}{MNIST\_Rule2}& &\multicolumn{3}{c}{MNIST\_Rule3}\\
 \cmidrule{2-4}
 \cmidrule{6-8}
 \cmidrule{10-12}
 $N$ & Baseline & Proposed & Gain & & Baseline & Proposed & Gain & & Baseline & Proposed & Gain\\
 \midrule
500 & 29.6 & 33.9  &14.5 & & 30.4 & 54.0 &77.6 & & 64.4 & 71.8  &11.5 \\ 
1000 & 81.0 & 83.2 &2.7  & & 83.9 & 86.5 &3.1 & & 82.8 & 84.8  &2.4\\
1500 & 84.8 & 87.8 &3.5  & & 88.8 & 90.8 &2.3 & & 89.6 & 90.2  &0.7\\
2000 & 89.4 & 91.0 &1.8  & & 89.6 & 94.3 & 5.2 & & 90.0 & 93.2  &3.6\\
 \bottomrule
  & & & & & & & & & & & \\
  \toprule
 \multirow{1}{*}{} & \multicolumn{3}{c}{SVHN\_Rule\,1} & & \multicolumn{3}{c}{SVHN\_Rule2}& &\multicolumn{3}{c}{SVHN\_Rule3}\\
 \cmidrule{2-4}
 \cmidrule{6-8}
 \cmidrule{10-12}
 $N$ & Baseline & Proposed & Gain & & Baseline & Proposed & Gain & & Baseline & Proposed & Gain\\
 \midrule
500 & 0.1 & 0.1 & 0  & & 0 & 0.2 &- & & 0 & 0 & - \\ 
1000 & 6.0 & 6.5 & 5.5  & & 9.0 & 9.4 &4.4  & & 14.6 & 15.9 & 8.9 \\
1500 & 17.4 & 20.2 & 16.1  & & 23.6 & 31.2 & 30.2 & & 15.2 & 20.2 & 32.9 \\
2000 & 22.1 & 25.9 & 17.2  & & 31.4 & 38.6 & 22.9 & & 28.2 & 39.8 & 41.1 \\
 \bottomrule
\end{tabular}
\end{table*}

\subsection{Missing Image with Blockout Data}
In real-life applications, it is common to have missing digits, for instance due to occlusions or noise. In some cases, a model aware of the underlying rule may be able to fill in these missing digits.

\begin{figure}[ht!]
\includegraphics[width=\linewidth]{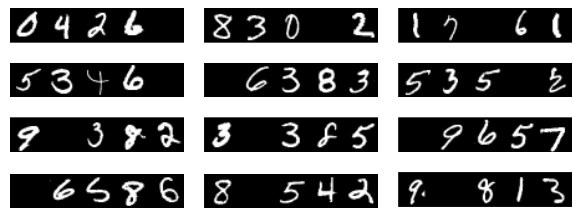}
\centering
\caption{Examples of Blockout images.}
\label{fig:blockout}
\end{figure}

We simulate this scenario by randomly blocking out a digit. We modify the test set images for three rules on MNIST, as illustrated in Fig. \ref{fig:blockout}. The results of our model on this modified dataset are shown in Table \ref{tab:blockout}. 
We see that, in most cases, adding the rule-based reinforcement term improves the accuracy. Moreover, we find that adding this term results in a relatively smaller loss in accuracy compared with original images. This suggests that incorporating rules can increase to some extent the model's robustness against missing numbers.

\begin{table*}[h]
    \centering
    \caption{Accuracy (\%) on Blockout datasets}
    \label{tab:blockout}
\begin{tabular}{ M{2.2cm}M{1.8cm}M{1.8cm}M{1.8cm}M{0.1cm}M{1.8cm}M{1.8cm}M{1.8cm}  }
\toprule
\multirow{1}{*}{} & \multicolumn{3}{c}{Baseline} & & \multicolumn{3}{c}{Proposed}\\
 \cmidrule{2-4}
 \cmidrule{6-8}
  & Original & Blockout & Gain & & Original & Blockout & Gain\\
\midrule
 MNIST\_Rule\,1 & 89.4 & 8.0 & -91.1 & & 91.0 &9.0 & \textbf{-90.1} \\ 
 MNIST\_Rule2 & 91.3 &14.0 & -84.7 & &91.6 &14.0 & -84.7\\ 
 MNIST\_Rule3 & 90.0 & 8.0 & -91.1 & & 92.2 &11.0 & \textbf{-88.1} \\ 
\bottomrule
\end{tabular}
\end{table*}

\subsection{Hard Digits}

Another source of errors can come from poorly written (e.g., slanted or incomplete) digits. Once more, by restricting possible digit sequences with given rules, the model may be able to guess the correct digit. We design the following experiment to simulate this scenario.

\begin{figure}[ht!]
\includegraphics[width=\linewidth]{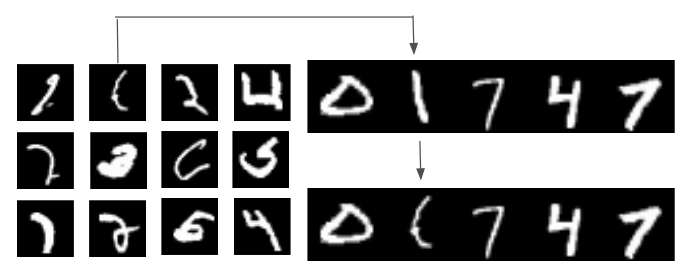}
\centering
\caption{Illustration of the procedure to generate hard samples.}
\label{fig:hard1}
\end{figure}

\begin{figure}[h]
\includegraphics[width=\linewidth]{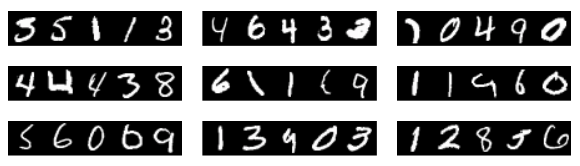}
\centering
\caption{Examples of hard samples.}
\label{fig:hard2}
\end{figure}

We train an image classifier with ResNet50 \cite{he2016deep} to classify MNIST digits. We early stop it so that the test accuracy is around 98.5\%. We then locate the test set images that are incorrectly classified. We call these set of images \textit{hard digits} (see Fig.~\ref{fig:hard1}). Afterwards, we modify the test datasets by randomly replacing a single digit with a hard digit one correspondong to the same value. We modify the 500 test images for each of the 3 rules on MNIST, as shown in Fig.~\ref{fig:hard2}. Results of this experiment can be found in Table \ref{tab:hardsample}. Once again, we see that incorporating the proposed rule-bases reinforcement in training can improve the model's robustness against hard digits, resulting in a less severe performance drop.  

\begin{table*}[ht!]
    \centering
    \caption{Accuracy(\%) on hard digits datasets.}
    \label{tab:hardsample}
\begin{tabular}{ M{2.2cm}M{1.8cm}M{1.8cm}M{1.8cm}M{0.1cm}M{1.8cm}M{1.8cm}M{1.8cm}  }
\toprule
\multirow{1}{*}{} & \multicolumn{3}{c}{Baseline} & & \multicolumn{3}{c}{Proposed}\\
 \cmidrule{2-4}
 \cmidrule{6-8}
  & Original & Blockout & Gain & & Original & Blockout & Gain\\
\midrule
 MNIST\_Rule\,1 & 89.4 & 33.0 & -63.9 & & 91.0 & 40.0 & \textbf{-56.0} \\ 
 MNIST\_Rule2 & 91.3 & 45.0 & -50.7 & &91.6 & 63.0 & \textbf{-31.2}\\ 
 MNIST\_Rule3 & 90.0 & 52.0 & -42.2 & & 92.2 & 59.0 & \textbf{-36.0} \\ 
\bottomrule
\end{tabular}
\end{table*}

\section{Discussion}

To demonstrate the usefulness of the proposed method, we evaluated it on a small synthetic dataset. To further validate our method, we aim to collect data from real applications like container number detection. A large-scale real-life dataset could be more challenging, however the results can be more transferable to practice. While this work focuses on digit sequence recognition, our method can be easily transferred to other recognition tasks as long as well-defined rules exist for target patterns. 

In future work, we would like to try more advanced algorithms like actor-critic \cite{konda2000actor} to reduce the gradient variance, or TRPO \cite{schulman2015trust} to increase the stability of each update. Value-based algorithms are another promising direction to improve results. For example, we can use the sum of logits as an approximation of Q value and perform Q-Learning \cite{mnih2013playing}. 
Another family of algorithms which could be investigated is oracle regression. In the current framework, we treat rules as a black-box oracle. However, we could also employ a neural net to imitate the behavior of the oracle, and we use the gradient of this neural network to increase the rule matching score \cite{foster2018practical}. 

\section{Conclusion}

In this work, we studied the problem of digit sequence recognition with non-differentiable rules. We synthesized 6 datasets from MNIST and SVHN for digit recognition tasks, using rules rooting from real-life applications. To exploit these rules, we advocate employed a reinforcement learning approach based on the policy gradient algorithm. Using a CRNN as baseline model, our experiments show that the model's accuracy can be improved in a variety of scenarios, including limited data and missing/indistinguishable numbers digits, when the proposed ruled-based reinforcement term is used during training.

\bibliography{biblio.bib}
\bibliographystyle{aaai}

\end{document}